\documentclass[pdflatex,sn-mathphys-num]{sn-jnl}
\usepackage{graphicx}%
\usepackage{multirow}%
\usepackage{amsmath,amssymb,amsfonts}%
\usepackage{amsthm}%
\usepackage{mathrsfs}%
\usepackage[title]{appendix}%
\usepackage{xcolor}%
\usepackage{braket}
\usepackage{qcircuit}
\usepackage{braket}
\usepackage{textcomp}%
\usepackage{manyfoot}%
\usepackage{booktabs}%
\usepackage{algorithm}%
\usepackage{algorithmicx}%
\usepackage{algpseudocode}%
\usepackage{listings}%
\usepackage{orcidlink}
\usepackage{type1cm}
\usepackage{mathrsfs}
\usepackage{subcaption}
\newtheorem{theorem}{Theorem}[section]
\newtheorem{corollary}[theorem]{Corollary}
\newtheorem{lemma}[theorem]{Lemma}

\begin{document}

\title[Data-Dependent Generalization Bounds for Parameterized Quantum Models Under Noise]{Data-Dependent Generalization Bounds for Parameterized Quantum Models Under Noise}

\author{\fnm{Bikram} \sur{Khanal}\orcidlink{0000-0003-2292-520X}}\email{bikram\_khanal1@baylor.edu}
\author*{\fnm{Pablo} \sur{Rivas}\orcidlink{0000-0002-8690-0987}}\email{pablo\_rivas@baylor.edu}
\affil{\orgdiv{School of Engineering \& Computer Science}, \orgname{Baylor University}, \orgaddress{\street{One Bear Place}, \city{Waco}, \postcode{76798}, \state{TX}, \country{USA}}}

\abstract{
Quantum machine learning offers a transformative approach to solving complex problems, but the inherent noise hinders its practical implementation in near-term quantum devices. This obstacle makes it difficult to understand the generalizability of quantum circuit models. Designing robust quantum machine learning models under noise requires a principled understanding of complexity and generalization, extending beyond classical capacity measures. This study investigates the generalization properties of parameterized quantum machine learning models under the influence of noise. We present a data-dependent generalization bound grounded in the quantum Fisher information matrix. We leverage statistical learning theory to relate the parameter space volumes and training sizes to estimate the generalization capability of the trained model. We provide a structured characterization of complexity in quantum models by integrating local parameter neighborhoods and effective dimensions defined through quantum Fisher information matrix eigenvalues. We also analyze the tightness of the bound and discuss the tradeoff between model expressiveness and generalization performance.
}


\maketitle

\section{Introduction}
Quantum machine learning (QML) holds the promise of revolutionizing data analysis by leveraging quantum mechanical principles to enhance computational efficiency and model expressivity~\cite{schuld2021machine,biamonte2017quantum,wang2024comprehensive}. Although still in its early stages~\cite{preskill2018quantum,torlai2020machine,khanal2024generalization}, QML has shown potential advantages over classical machine learning in certain tasks, such as classification~\cite{havlivcek2019supervised,cerezo2021variational,rivas2024Quanvolutional}, regression~\cite{lloyd2013quantum}, and clustering~\cite{aimeur2007quantum}. A crucial aspect of any machine learning model, including QML models, is its generalization ability. Generalization refers to a model's ability to perform well on unseen data~\cite{mohri2018foundations,abu2012learning,shalev2014understanding}. Generalization bounds provide theoretical guarantees on the performance of a model in new data and play a vital role in ensuring the reliability and robustness of machine learning models~\cite{emami2020generalization,jakubovitz2019generalization,nadeau1999inference}.

One of the prominent questions in quantum machine learning is understanding how noise affects the generalizability of models~\cite{banchi2021generalization,gil2023understanding,caro2021encoding}. Generalization bounds are typically derived by bounding a complexity measure of the hypothesis class used for learning~\cite{caro2021encoding}; such measures—like Rademacher complexity, VC dimension, and metric entropy—quantify the expressiveness or capacity of the model class. These bounds are especially crucial in the NISQ era~\cite{khanal2023noise,caro2022generalization,khanal2024generalization,caro2021encoding,caro2024information,gil2023understanding}, where devices are characterized by limited qubit coherence times, suboptimal gate fidelities, and a high susceptibility to noise, all of which can significantly impact the performance and reliability of QML models. Several studies have investigated QML generalization bounds—exploring approaches such as uniform bounds, margin-based bounds, and stability analysis~\cite{caro2022generalization,haug2023generalization,banchi2021generalization,khanal2024generalization,jakubovitz2019generalization,caro2024information,gil2023understanding,caro2021encoding,canatar2022bandwidth,caro2023out,abbas2021power,bu2023effects,bu2022statistical}. Recent works by Bu et al.\cite{bu2022statistical} have further refined this framework by rigorously linking the complexity of QML models to their quantum resource content through tailored Rademacher and Gaussian analyses. In particular, they introduce a $(p,q)$-norm to quantify the non-stabilizer (or “magic”) resource, demonstrating that the incorporation of non-Clifford gates—characterized via free robustness measures—directly influences the generalization capacity of quantum circuits\cite{bu2023effects}. These contributions offer valuable insights into how circuit architecture, data encoding strategies, and optimization choices affect QML generalization. Concurrently, research has also examined the broader impact of noise on QML~\cite{preskill2018quantum,martinis2003decoherence,wang2021noise,heyraud2022noisy}, emphasizing that noise from sources such as decoherence~\cite{shor1995scheme}, gate errors, and other quantum phenomena poses formidable challenges to developing models that can reliably learn and generalize. Although several strategies for mitigating noise have been proposed~\cite{khanal2024learning,shaib2023efficient,ferracin2024efficiently,shaib2023efficient}, much of the existing research on QML generalization bounds focuses on idealized, noise-free quantum computers. Moreover, while the generalization capabilities of quantum neural networks and quantum kernel methods have been explored~\cite{gentinetta2024complexity,liu2021rigorous,kubler2021inductive}, these bounds are often tailored to specific algorithmic techniques or problems. Some studies~\cite{wang2021noise,thanasilp2024exponential,preskill2018quantum,huang_2021,kubler2021inductive,heyraud2022noisy} further suggest that quantum noisy channels can significantly hinder QML capabilities. We note that the generalization abilities of quantum neural network models may also be compromised by the barren plateau phenomenon~\cite{czarnik2021error,cerezo2022challenges,holmes2022connecting,zhao2021analyzing,wang2021noise,mcclean2018barren,arrasmith2021effect} and that quantum kernel methods can suffer from exponential concentration~\cite{thanasilp2024exponential}. It should be noted that training-induced noise differs from the inherent noise in quantum devices and is not considered in this work.  Consequently, a comprehensive understanding of how noise affects the generalization performance of QML models remains lacking—a gap that is particularly critical in the NISQ era, where noise is unavoidable and can severely limit the practical applicability of QML algorithms.

This work addresses this gap by investigating the generalization bound for QML under a noisy channel. Specifically, we consider a scenario where a supervised QML model is trained on a noisy quantum system, and the goal is to establish a bound on its generalization error in the presence of noise. We achieve this goal by leveraging the quantum Fisher information-derived metrics, parameter space volume, and training sample size into a quantifiable measure of complexity in noisy QML models and derive the generalization bound for near-term QML models. Our approach interprets complexity as a controllable quantity rather than an abstract or unbounded factor. Specifically, we consider depolarization noise channels for an experiment to investigate how this noise affects both the theoretical generalization bound and the empirical performance of the QML models. The work refines the global bound by considering local parameter neighborhoods, effective dimensions defined through quantum Fisher information matrix eigenvalues, and conditions that stabilize complexity growth under noise. The statistical results verify that the proposed local refinements for the depolarization noise channel provide a more accurate and efficient analysis of the generalization performance. 
Consequently, our work bridges abstract theoretical concepts and the operational needs of quantum learning architectures. The contribution lies not in proposing new computational primitives but in producing a rigorous theoretical structure that uses geometric principles to inform model design. Rather than treating all parameters equally, our approach clarifies that parameter directions differ in their contribution to predictive performance, making geometric considerations not just aesthetic choices but practical necessities.

The remainder of this paper proceeds as follows. The next section presents a preliminary background in machine learning and the quantum machine with the formal setup of binary classification learning problems. We briefly discuss classical and quantum Fisher information in section~\ref{sec:Fisher}, and the effective dimension in section~\ref{sec:effectivedimension}. Section~\ref{sec:generalizationbound} proposes the generalization bound and corollaries followed by a local refinement. A subsequent section~\ref{sec:numericalAnalysis} discusses empirical validation through numerical experiments, confirming the agreement with theoretical predictions. We discuss these results in a broader context in section~\ref{sec:discussion} and summarize key contributions in section~\ref{sec:conclusion}. Appendix~\ref{Appendix} provides detail in the mathematical derivation of the proposed bound with the necessary background.

\section{Preliminary Definitions }\label{sec:preliminary}
We begin by discussing the fundamental concepts of learning from data and extending them to the QML framework within a supervised learning setting. Readers already familiar with the field may skip the following two subsections. For those seeking a deeper understanding, we recommend the books by~\cite{shalev2014understanding,abu2012learning} for classical learning theory and the books by~\cite{schuld2021machine,nielsen2002quantum} for QML and quantum information background.

\subsection{Supervised Learning}
In supervised learning, we are given the input space $\mathcal{X}$ and the output space $\mathcal{Y}$. Let $P$ be the joint probability distribution over $\mathcal{X} \times \mathcal{Y}$ and $D = \{(x_i, y_i)\}_{i=1}^N$ be a dataset of $N$ i.i.d samples drawn from $P$. Supervised learning aims to learn a function $f: \mathcal{X} \rightarrow \mathcal{Y}$ that maps input to output. Given a training dataset $D$, the learning algorithm $\mathcal{A}$ selects the best function $h$ from a hypothesis class $\mathcal{H}$. Usually, the elements of $\mathcal{H}$ are parameterized by a vector of parameters $\theta \in \Theta$, where $\Theta  \subset \mathbb{R}^d$ is a parameter space of $d\in \mathbb{N}$ dimension. We may also refer to $h$ as $h_\theta$ to emphasize the dependence on the parameters and define the hypothesis class as $\mathcal{H} = \{h_\theta: \theta \in \mathbb{R}^d\}$. Please note that  $h$ and $f$ are used interchangeably as hypotheses in the context of learning, and this might be exhibited in our manuscript, too.

Given two approximations $h_1,h_2 \in \mathcal{H}$, how do we decide which is better? One way to do that is to define a loss function $l: \mathcal{Y} \times \mathcal{Y} \rightarrow \mathbb{R}$. The loss function measures the difference between the predicted output $\hat{y} \in \mathbb{R}$ and the true output $ y \in {\mathcal{Y}}$. The learning algorithm aims to find $h$ that minimizes the expected loss (or expected risk) defined as:
\begin{equation}\label{eq:expectedRisk}
    R(h) = \mathbb{E}_{(x,y) \sim P}[l(y,h(x))].
\end{equation}
The expected risk is the average loss over all possible inputs and outputs. Therefore, the goal is to find a function $h$ such that $l(h)$ is as small as possible. In practice, we cannot access the true distribution $P$; hence, it is often challenging to compute the expected risk. Instead, we compute the empirical risk and try to minimize it, also referred to as the empirical risk minimization (ERM) principle. The empirical risk is defined as:
\begin{equation}\label{eq:empiricalRisk}
    \hat{R}(h) = \frac{1}{N} \sum_{i=1}^N l(y_i,h(x_i)).
\end{equation}

A central goal of machine learning is to understand how well the empirical risk approximates the expected risk~\cite{shalev2010learnability,mohri2018foundations,shalev2014understanding,abu2012learning}. The difference between the empirical and expected risks is the generalization error. The generalization error quantifies how well the model generalizes from the training data to unseen data. A model with low generalization error is said to generalize well, while a model with high generalization error suggests poor generalization. Generalization error is typically of the form:
\begin{equation}\label{eq:generalizationError}
    \text{GenError} = R(h) - \hat{R}(h).
\end{equation}
One of the important uses of generalization error is to provide a bound on the expected risk of the model. We can use the probability to bound the generalization error as:
\begin{equation} \label{eq:converganceBound}
    \text{Pr}\left[ \sup \limits_{h \in \mathcal{H}} \left| \hat{R}(h) - R(h)\right| \leq \epsilon \right] \geq 1 - \delta,
\end{equation}
where $\text{Pr}[\cdot]$ denotes the probability measure, \textit{sup} is the supremum, $\epsilon$ is the error tolerance, and $\delta$ is the confidence level. The uniform convergence bound given by Eq.~\eqref{eq:converganceBound} implies that with probability at least $1 - \delta$, the empirical risk of the best hypothesis in the hypothesis class is close to the expected risk. We can use the uniform convergence bound approach to provide a bound on the generalization error, i.e., generalization bound. The generalization bound guarantees how well the model generalizes to unseen data~\cite{valle2020generalization,johansson2022generalization}. Generalization bound is typically of the form:
\begin{equation}\label{eq:complexityBound}
    R(h) \leq \hat{R}(h) + \text{complexity term},
\end{equation}
where the complexity term is often a function of the $\mathcal{H}$ and $N$. Complexity measure depends on $P$ over $\mathcal{X} \times \mathcal{Y}$, and if $P$ is easy to learn, the complexity term will be small. In other words, if we are to find the model with the low complexity term, we can expect the model to generalize well~\cite{pape2015complexity,bialek2001predictability}.

Alternatively, instead of getting a bound on the probability of the generalization error, we can obtain an upper bound in generalization error expectations instead. i.e., we can bound the expected generalization error as:
\begin{equation}\label{eq:expectedcomplexityBound}
    \mathbb{E}_{D} \left[ \sup \limits_{h \in \mathcal{H}} \left| \hat{R}(h) - R(h)\right| \right]  \leq \text{complexity term},
\end{equation}
Where the expectation is taken over the training dataset $D$. In this work, we focus on the expected generalization error bound and provide a bound on the expected generalization error for quantum machine learning models in noisy environments. Before proceeding, we briefly discuss the quantum machine learning framework.

\subsection{Quantum Machine Learning}\label{sec:quantum_machine_learning}
Unlike classical ML, which relies on deterministic computations, QML leverages the inherently probabilistic nature of quantum mechanics, enabling it to explore multiple states simultaneously and potentially solve problems that are intractable classically. This property allows QML algorithms to operate in high-dimensional Hilbert spaces with notable efficiency~\cite{schuld2019quantum,havlivcek2019supervised}. The central idea in QML for the near-term era is to encode classical data into quantum states, process them through quantum circuits, and extract predictions through measurements. The quantum circuit is typically parameterized by a set of angles, which are optimized to minimize a chosen cost function. The optimization process is often performed using gradient-based methods, where the gradients are computed using the parameter-shift rule~\cite{mitarai2018quantum,schuld2019evaluating}. Like classical ML models, the quantum circuit model is trained to minimize the empirical risk. The key difference is that the quantum circuit model operates on quantum states and measurements, which can lead to quantum advantages in certain tasks~\cite{havlivcek2019supervised,cerezo2021variational}. The generalization properties of QML models are particularly interesting, as they can provide insights into the model's performance on unseen data.

We consider a binary classification setting with a classical input space $\mathcal{X}\subset \mathbb{R}^m$ and a label space $\mathcal{Y}=\{-1,+1\}$. Let $\mathcal{H}_s$ be a Hilbert space associated with an $n$-qubit quantum system, so $\dim(\mathcal{H}_s)=2^n$. A central goal is to encode each classical data point $x\in\mathcal{X}$ into a quantum state and process it through a parameterized quantum circuit (PQC) to obtain measurement outcomes that can be interpreted as predictions.

\subsubsection{Data Encoding}\label{subsubsec:data_encoding}
A key step in quantum machine learning is encoding the classical data $x$ into a quantum state. Define a feature map 
\(
\phi:\mathcal{X}\rightarrow \mathcal{H}_s
\)
that maps each $x\in \mathcal{X}$ to a normalized quantum state $\ket{\phi(x)} \in \mathcal{H}_s$. This encoding can be achieved in various ways, depending on the hardware constraints and the nature of the data~\cite{schuld2019quantum,biamonte2017quantum}. Ref.~\cite{schuld2021effect} is an excellent resource for understanding data encoding strategies in quantum machine learning. A common strategy is to apply parameterized single-qubit rotations whose angles depend on the components of $x$:
\[
R_x(x_i) = \exp(-i x_i \sigma_x /2),\quad R_z(x_i) = \exp(-i x_i \sigma_z /2), \quad R_y(x_i) = \exp(-i x_i \sigma_y /2),
\]
where $\sigma_x, \sigma_y \text{ and } \sigma_z$ are Pauli matrices. These operations can be arranged so that the $m$ features of $x$ are distributed across $n_{\phi} \le n$ qubits, resulting in the encoded state $\ket{\phi(x)}$ defined as:
\begin{equation}\label{eq:initialstate_new}
\ket{\psi(x)} \;=\; \ket{\phi(x)} \;\otimes\; \ket{0}^{\otimes (n - n_{\phi})}.
\end{equation}

\subsubsection{Parameterized Quantum Circuit}
Following data encoding, we apply a PQC with trainable parameters 
\[
\theta = \bigl(\theta_1,\theta_2,\dots,\theta_L\bigr).
\]
Each $\theta_k$ can represent the rotation angles in single or multi-qubit gates. Mathematically, the PQC is a unitary $U(\theta)$ that is typically decomposed into layers of parameterized gates and entangling (non-parameterized) gates:
\[
U(\theta) \;=\; \prod_{k=1}^L U_k(\theta_k),
\]
Where $U_k(\theta_k)$ often acts on one or more qubits. Acting on the initial state \(\ket{\psi(x)}\), the PQC produces
\begin{equation}\label{eq:PQC_new}
\ket{\psi_\theta(x)} \;=\; U(\theta)\,\ket{\psi(x)},
\end{equation}
whose density matrix representation is
\begin{equation}\label{eq:densitymatrix_new}
\rho_\theta(x) \;=\; \ket{\psi_\theta(x)}\bra{\psi_\theta(x)}.
\end{equation}

In quantum machine learning, without loss of generalizability, the quantum neural networks~\cite{benedetti2019parameterized,cerezo2021variational,gyongyosi2019training} can also be considered PQC. The analogy to classical neural networks arises because the PQC parameters $\theta$ are iteratively optimized (trained) to minimize a chosen cost function. While backpropagation is central to classical networks, the gradient can be computed via specialized quantum techniques, such as the parameter-shift rule (detailed in Section~\ref{subsubsec:parameter_shift_rule}).

\subsubsection{Measurement and Classification}
To extract a classical prediction, we measure a designated subset of qubits. In a simple binary classification scenario, we measure only the first qubit with a two-outcome measurement described by a positive operator-valued measure (POVM) $\{M_{+1}, M_{-1}\}$:
\[
M_{+1} \;=\; \ket{0}\bra{0}\;\otimes\; I^{\otimes (n-1)}, 
\quad 
M_{-1} \;=\; \ket{1}\bra{1}\;\otimes\; I^{\otimes (n-1)}.
\]
where $\ket{0} = \begin{bmatrix}
1 \\
0
\end{bmatrix}, \bra{0} =\ket{0}^{\dag} \text{ and } I$ is an identity matrix. The probability of obtaining label $y\in\{+1,-1\}$ for input $x$ is then:
\begin{equation}\label{eq:probability_new}
\hat{y} = p(y \mid x;\theta) \;=\; \mathrm{Tr}\!\bigl[M_y\,\rho_\theta(x)\bigr].
\end{equation}
A simple classification rule predicts $y=+1$ if $\hat{y}\ge 0.5$ and $y=-1$ otherwise, mirroring classical threshold-based decision rules.

\subsubsection{Training and Optimization}\label{subsubsec:training_optimization}
In supervised quantum machine learning, we define a loss function (cost function) $\mathcal{L}(\theta)$ that quantifies prediction error across a training set $\{(x_i,y_i)\}$. A typical example is the cross-entropy loss:
\begin{equation}
\mathcal{L}(\theta) \;=\; -\frac{1}{N}\sum_{i=1}^N \Bigl[ y_i \log (\hat{y_i}) + (1-y_i) \log (1-\hat{y})\Bigr].
\end{equation}
We then seek optimal parameters $\theta^*$ by minimizing $\mathcal{L}(\theta)$ via gradient-based methods:
\begin{equation}
\theta \;\leftarrow\; \theta \;-\;\alpha\,\nabla_\theta \mathcal{L}(\theta),
\end{equation}
Where $\alpha$ is a learning rate. Since the entire computation of $p(y\mid x;\theta)$ is performed by a quantum circuit, we need a way to obtain $\nabla_\theta \mathcal{L}(\theta)$ from measurement statistics.

\subsubsection{Parameter-Shift Rule for Gradient Computation}\label{subsubsec:parameter_shift_rule}
A popular approach for gradient computation in near-term quantum devices is the \emph{parameter-shift rule}~\cite{mitarai2018quantum,schuld2019evaluating,crooks2019gradients}. Consider a circuit with a parameterized gate of the form:
\[
U(\theta_j) \;=\; \exp\!\bigl(-i\,\theta_j\,G/2\bigr),
\]
where $G^2 = I$ (common for single-qubit Pauli rotations). For an observable $M$ measured at the end of the circuit, the parameter-shift rule states that:
\begin{equation}\label{eq:param_shift_general}
\frac{\partial}{\partial \theta_j} \langle M\rangle(\theta) 
\;=\; \frac{1}{2}\,\Bigl[\langle M\rangle\bigl(\theta_j + \tfrac{\pi}{2}\bigr) - \langle M\rangle\bigl(\theta_j - \tfrac{\pi}{2}\bigr)\Bigr],
\end{equation}
where $\langle M\rangle(\theta) = \mathrm{Tr}[M\,\rho_\theta(x)]$ is the expectation value of $M$ (or an effective estimator for $p(y\mid x;\theta)$). This formula allows one to compute partial derivatives by running the circuit twice, at shifted parameter values $\theta_j \pm \frac{\pi}{2}$. In practice, one repeats these measurements multiple times to estimate the required probabilities, then aggregates these partial derivatives across all parameters to form $\nabla_\theta \mathcal{L}(\theta)$.

\subsubsection{Practical Considerations and Noise}
Current quantum hardware (NISQ devices) are subject to noise and decoherence, which can degrade performance. In the presence of a noisy quantum channel $\mathcal{N}$, the noiseless density operator $\rho_\theta(x)$ transforms into 
\begin{equation}
\tilde{\rho}_\theta(x) \;=\; \mathcal{N}\bigl(\rho_\theta(x)\bigr).
\end{equation}
As a result, the measured probabilities become
\begin{equation}\label{eq:noisyProbability}
\tilde{p}(y\mid x;\theta) \;=\; \mathrm{Tr}\bigl[M_y\, \tilde{\rho}_\theta(x)\bigr]
\;=\;\mathrm{Tr}\Bigl[M_y\, \mathcal{N}(\rho_\theta(x))\Bigr].
\end{equation}

For certain noise channels and measurement bases, this noisy probability $\tilde{p}(y|x;\theta)$ may be well-approximated by a differentiable function $\eta:[0,1]\to (0,1]$ of the noiseless probability \(\eta\bigl(p(y\mid x;\theta)\bigr)\) as:
\begin{equation}\label{eq:noisyFunction}
\tilde{p}(y|x;\theta) = \eta\left(p(y|x;\theta)\right),
\end{equation}
with $\eta$ continuous in the noise rate and $\eta(0)=1$~\cite{holevo2019quantum,guctua2006local}. This simplification holds, for example, when the noise operators commute with the measurement operators, as is the case with a depolarizing channel that acts on states measured on the computational basis~\cite{magesan2011scalable,preskill1998lecture,nielsen2002quantum,khanal2024modified}.

Noise modifies the sensitivity of the output probability distribution to changes in $\theta$, influencing both trainability (e.g., risk of barren plateaus~\cite{mcclean2018barren,cerezo2021variational}) and generalization~\cite{haug2024generalization}. Despite challenges related to data encoding overhead, hardware noise, and the limitations of near-term devices, PQC-based models remain promising candidates for quantum advantage in specific domains~\cite{gyongyosi2019training,gyongyosi2020circuit,gyongyosi2021scalable}. Because noise modifies the sensitivity of output probabilities with respect to variations in $\theta$, analyzing $\tilde{p}(y|x;\theta)$ under realistic noise channels is crucial for understanding trainability and generalization. This sensitivity can be quantified through the Fisher information matrices, both classical and quantum, which serve as fundamental tools for analyzing parameter distinguishability and generalization in the presence of noise~\cite{haug2024generalization,kasatkin2024detecting,bharti2021fisher,meyer2021fisher}.

\section{Fisher Information}\label{sec:Fisher}
From a machine learning perspective, the Fisher information measures the sensitivity of the output label distribution with respect to changes in model parameters~\cite{amari1998natural,haug2021capacity,martens2020new,stokes2020quantum}. Higher Fisher information indicates that the parameters $\theta$ carry more information about the output label $ y$, enabling more precise parameter estimation and potentially improving learnability. In a quantum machine learning setting, we must consider classical and quantum notions of Fisher information to capture the role of quantum states and measurements fully.

\subsection{Classical Fisher Information}\label{subsec:CFI}

Consider a parameterized probability distribution $p(y|x;\theta)$, where $y\in\mathcal{Y}$ denotes the label and $\theta=(\theta_1,\dots,\theta_d)$ are the model parameters. The classical Fisher information matrix (FIM) is defined as:
\begin{equation}\label{eq:FisherInformation}
\mathcal{F}(\theta)_{ij} = \sum\limits_{y \in \mathcal{Y}} p(y|x;\theta) \left( \frac{\partial \ln p(y|x;\theta)}{\partial \theta_i} \right)\left( \frac{\partial \ln p(y|x;\theta)}{\partial \theta_j} \right).
\end{equation}
Equivalently, this can be written as:
\begin{equation}\label{eq:FisherInformationSimplified}
\mathcal{F}(\theta)_{ij} = \sum\limits_{y \in \mathcal{Y}} \frac{1}{p(y|x;\theta)} \left( \frac{\partial p(y|x;\theta)}{\partial \theta_i} \right)\left( \frac{\partial p(y|x;\theta)}{\partial \theta_j} \right).
\end{equation}

The classical FIM quantifies how sensitive the observed label distribution is to small perturbations of the parameters. PQC measurement outcomes form the probability distribution. However, in the near-term there is the exponential growth of potential measurement outcomes as the number of quantum bits increases~\cite{baumgratz2013scalable}. This means many measurement outcomes might never appear, although with minimal probability, causing instability in classical Fisher information matrix computation~\cite{haug2021capacity,meyer2021fisher}.

\subsubsection{Fisher Information Under a Noisy Channel}\label{subsubsec:CFIunderNoise}
When a noisy quantum channel $\mathcal{N}$ acts on the underlying quantum state, the probability distribution $p(y|x;\theta)$ is replaced by $\tilde{p}(y|x;\theta)$ defined in Eq.~\eqref{eq:noisyProbability}. The corresponding noisy classical FIM becomes:
\begin{equation}\label{eq:FisherInformationNoisy}
\tilde{\mathcal{F}}(\theta)_{ij} = \sum\limits_{y \in \mathcal{Y}} \frac{1}{\tilde{p}(y|x;\theta)} \left( \frac{\partial \tilde{p}(y|x;\theta)}{\partial \theta_i} \right)\left( \frac{\partial \tilde{p}(y|x;\theta)}{\partial \theta_j} \right).
\end{equation}

Using the relationship in Eq.\eqref{eq:noisyFunction} and the chain rule, we have:
\begin{equation}\label{eq:noisyexpderivative}
\frac{\partial \tilde{p}(y|x;\theta)}{\partial \theta_i} = \eta^\prime\left(p(y|x;\theta)\right) \frac{\partial p(y|x;\theta)}{\partial \theta_i},
\end{equation}
where $\eta$ and $\eta^{\prime}$ characterize the noise-induced transformation of the probabilities. Since $\eta$ is differentiable, $\eta^{\prime}$ can be approximated by the derivative of $\eta$ at $p(y|x;\theta)$. The exact form of $\eta$ depends on the noise model and the measurement basis, but it is often chosen to be a smooth function that preserves the ordering of probabilities. Now, substituting Eq.\eqref{eq:noisyexpderivative} into Eq.~\eqref{eq:FisherInformationNoisy}, we get:
\begin{equation}\label{eq:FIMNoisy}
\tilde{\mathcal{F}}(\theta)_{ij} = \sum\limits_{y \in \mathcal{Y}} \frac{\left[\eta^\prime(p(y|x;\theta))\right]^2}{\eta(p(y|x;\theta))} \left( \frac{\partial p(y|x;\theta)}{\partial \theta_i}\right)\left(\frac{\partial p(y|x;\theta)}{\partial \theta_j}\right).
\end{equation}

This expression shows that the noisy classical FIM is the noiseless FIM scaled by factors depending on $\eta$ and $\eta^{\prime}$. If $\eta^{\prime}\approx 1$, noise minimally affects gradients. If $\eta^{\prime}>1$, the model is more sensitive to parameter changes, possibly improving short-term learnability but risking overfitting. If $\eta^{\prime}<1$, the parameter sensitivity is suppressed, potentially resulting in slower training and higher bias. These dynamics reflect the interplay between noise and trainability in quantum machine learning models.

The classical FIM considered so far treats measurement outcomes as classical data. While this is sufficient for fixed measurements, it does not exploit the full quantum nature of the underlying states. To fully characterize parameter sensitivity at the quantum level, one must turn to the quantum Fisher information, which provides a more fundamental and measurement-independent measure of distinguishability of quantum states~\cite{braunstein1994statistical,liu2020quantum,fujiwara2001quantum}.

\subsection{Quantum Fisher Information}\label{sub:QFI}

The quantum Fisher information (QFI)~\cite{liu2020quantum,meyer2021fisher,petz2011introduction,haug2021capacity} generalizes the concept of Fisher information to the quantum domain. Instead of fixing a particular measurement, the QFI considers the most informative measurement possible. This makes the QFI a fundamental limit on parameter precision achievable with an optimal choice of measurement.

\subsubsection{Ideal (Pure) States}

For a single-parameter family of pure states $|\psi(\theta)\rangle$, the QFI is given by~\cite{yamamoto2019natural,liu2014fidelity}:
\begin{equation}\label{eq:pure_qfi}
F_Q(\theta) = 4 \left[ \langle \partial_\theta \psi(\theta) | \partial_\theta \psi(\theta) \rangle - |\langle \psi(\theta) | \partial_\theta \psi(\theta) \rangle|^2 \right].
\end{equation}

If $|\psi(\theta)\rangle = e^{-iG\theta}|\psi(0)\rangle$ for some Hermitian generator G, this simplifies to:
\begin{equation}\label{eq:pure_qfi_gen}
F_Q(\theta) = 4 \langle \psi(0)|(\Delta G)^2|\psi(0)\rangle,
\end{equation}
where $(\Delta G)^2 = G^2 - \langle G\rangle^2$.

\subsubsection{Noisy (Mixed) States}\label{subsec:mixedfisher}

For a mixed state $\rho(\theta)$, the QFI is defined using the symmetric logarithmic derivative (SLD) $L_\theta$, which satisfies:
\begin{equation}\label{eq:sld_equation}
\frac{\partial \rho(\theta)}{\partial \theta} = \frac{1}{2} (L_\theta \rho(\theta) + \rho(\theta)L_\theta).
\end{equation}
The QFI is then:
\begin{equation}\label{eq:mixed_qfi}
F_Q(\theta) = \text{Tr}[\rho(\theta) L_\theta^2].
\end{equation}

If $\rho(\theta)=\sum_i \lambda_i(\theta)|\psi_i(\theta)\rangle\langle\psi_i(\theta)|$ is the spectral decomposition, following~\cite{liu2014fidelity} one can write:
\begin{equation}\label{eq:sld_spectral}
L_\theta = \sum_{i,j} \frac{2}{\lambda_i(\theta) + \lambda_j(\theta)} \langle \psi_i(\theta)|\partial_\theta \rho(\theta)|\psi_j(\theta)\rangle |\psi_i(\theta)\rangle\langle\psi_j(\theta)|.
\end{equation}

For multiple parameters $\theta=(\theta_1,\dots,\theta_d)$, the QFI generalizes to a matrix known as the quantum Fisher information matrix (QFIM). The QFIM's elements can be derived from the SLDs corresponding to each parameter, and it encodes the distinguishability of $\rho(\theta)$ in all parameter directions. This multi-parameter QFIM reduces to the single-parameter QFI if we focus on one parameter at a time.

\subsection{Relationship between Classical and Quantum Fisher Information}\label{subsec:CFIQFIrelationship}

Consider a measurement described by a POVM $\{M_x\}$, where the probability of obtaining outcome x is $p(x|\theta)=\text{Tr}[\rho(\theta)M_x]$. The corresponding classical Fisher information (CFI) is:
\begin{equation}
F_C(\theta) = \sum_x p(x|\theta)\left(\frac{\partial}{\partial \theta}\ln p(x|\theta)\right)^2.
\end{equation}

The Quantum Cramér-Rao bound states:
\begin{equation}\label{eq:qcramer_rao}
\mathrm{Var}(\hat{\theta}) \geq \frac{1}{M F_Q(\theta)},
\end{equation}
where $\hat{\theta}$ is an unbiased estimator, and $M$ is the number of experiments. Since the QFI represents the maximal information obtainable over all possible measurements, we have:
\begin{equation}\label{eq:cfi_qfi_inequality}
F_C(\theta) \leq F_Q(\theta),
\end{equation}
with equality when the chosen POVM attains the quantum limit.

This relationship shows that the QFI provides a fundamental upper bound on the precision achievable by any measurement scheme, making it a key tool in quantum parameter estimation~\cite{meyer2021fisher}. Classical Fisher information depends on the chosen measurement, whereas quantum Fisher information is measurement-independent and sets the ultimate precision limit.

Classical and quantum Fisher information concepts and their interplay form the foundation for defining complexity measures such as the effective dimension of a parameter space. The following section introduces the effective dimension and its role in understanding generalization in quantum machine learning models.

\section{Effective Dimension}\label{sec:effectivedimension}
The notion of effective dimension~\cite{abbas2021effective,schuld2021effect,haug2024generalization} offers a way to measure the intrinsic complexity of a parameterized quantum model beyond the raw parameter count. While  $d$ is the total number of parameters, many directions in this $d$-dimensional parameter space may contribute minimally to changes in the model's outputs. The effective dimension systematically determines which parameter directions substantially alter observable outcomes and which do not. This perspective is particularly beneficial for quantum models, where certain parameter variations may be "inert" due to noise, hardware constraints, or limited sensitivity of the state to those parameters.

Mathematically, a large parameter space can still exhibit an effectively smaller "active" subspace if specific parameters fail to induce meaningful changes in measurement statistics. Informally, one may imagine each parameter as contributing a "direction" along which the quantum state could change. If moving along some directions hardly affects the measurable quantities, the system effectively behaves as if those parameters do not exist. In practice, this phenomenon leads to an effective dimension $d_{\mathrm{eff}} \leq d$. Such a reduction in the dimension of relevant variability can mitigate overfitting in machine-learning contexts but also limit representational capacity if too many directions are inactive.

A central tool for quantifying parameter influence is the quantum Fisher information matrix. For a pure state \(\ket{\psi(\theta)}\), the QFIM is defined by,
\begin{equation}\label{eq:qfim}
F_{ij}(\theta) = 4\mathrm{Re}\left[\langle \partial_i \psi(\theta)|\partial_j \psi(\theta)\rangle - \langle \partial_i \psi(\theta)|\psi(\theta)\rangle \langle \psi(\theta)|\partial_j \psi(\theta)\rangle \right],
\end{equation}
where \(\partial_i \ket{\psi(\theta)}\) denotes the partial derivative of \(\ket{\psi(\theta)}\) with respect to the \(i\)-th parameter. Analogous formulations exist for mixed states (detailed in Appendix~\ref{subsec:mixedfisher}, see also \cite{meyer2021fisher}). The QFIM captures how sensitively the state responds to small variations in each parameter. Directions in parameter space that do not appreciably shift the state's geometry yield small or negligible eigenvalues in the QFIM, effectively indicating "inactive" parameters.

One way to formalize this idea is via the rank of the QFIM:
\begin{equation}\label{eq:deffrank}
d_{\mathrm{eff}} 
\;=\; 
\max_{\theta \,\in\, \Theta} \text{rank}\!\bigl(F(\theta)\bigr).
\end{equation}
If \( \text{rank}\!\bigl(F(\theta)\bigr)\) is less than \(d\), it signifies that some parameters do not induce linearly independent changes in the quantum state or its measurement outcomes, leading to $d_{\mathrm{eff}}<d$. Consequently, the manifold of states effectively spans fewer directions than the raw parameter count might suggest.

Rather than only checking the rank, one may study the QFIM's eigenvalues \(\{\lambda_i\}_{i=1}^d\). A common measure is:
\begin{equation}\label{eq:eigenvalue_effective_dimension}
d_{\mathrm{eff}} 
\;\approx\; 
\frac{\left(\sum_{i=1}^d \lambda_i\right)^2}{\sum_{i=1}^d \lambda_i^2},
\end{equation}
an expression closely related to the inverse participation ratio. Here, if the QFIM spectrum is sharply concentrated in a handful of large eigenvalues, $d_{\mathrm{eff}}$ remains well below $d$. Conversely, if many eigenvalues are comparable, the system effectively leverages a broader portion of parameter space.

Effective dimension has multifaceted importance in QML. For learning models, a reduced $d_{\mathrm{eff}}$ can provide greater robustness against overfitting since only a subset of parameters drives significant changes in the loss landscape, effectively limiting the complexity of the model class. A high $d_{\mathrm{eff}}$ implies that multiple parameters can be measured with high precision simultaneously. In contrast, a low $d_{\mathrm{eff}}$ suggests that sensitivity is concentrated along only a few principal directions. In both settings, awareness of the effective dimension can guide model selection, training strategies, or experimental design by highlighting how many "truly operative" degrees of freedom the quantum system has.

Thus, the effective dimension concept refines the naive view of dimensionality by honing in on the directions in which the model or system genuinely exhibits variability. By treating parameters that do not measurably alter outcomes as effectively inactive, $d_{\mathrm{eff}}$ offers a more precise view of capacity and complexity in quantum models. Combined with the QFIM framework, this facilitates theoretical analyses of generalization, performance, and practical insights into designing and training near-term quantum architectures for learning.

\section{Generalization bound}\label{sec:generalizationbound}
\begin{figure*}[ht]
    % \centering
    \includegraphics[width=\textwidth]{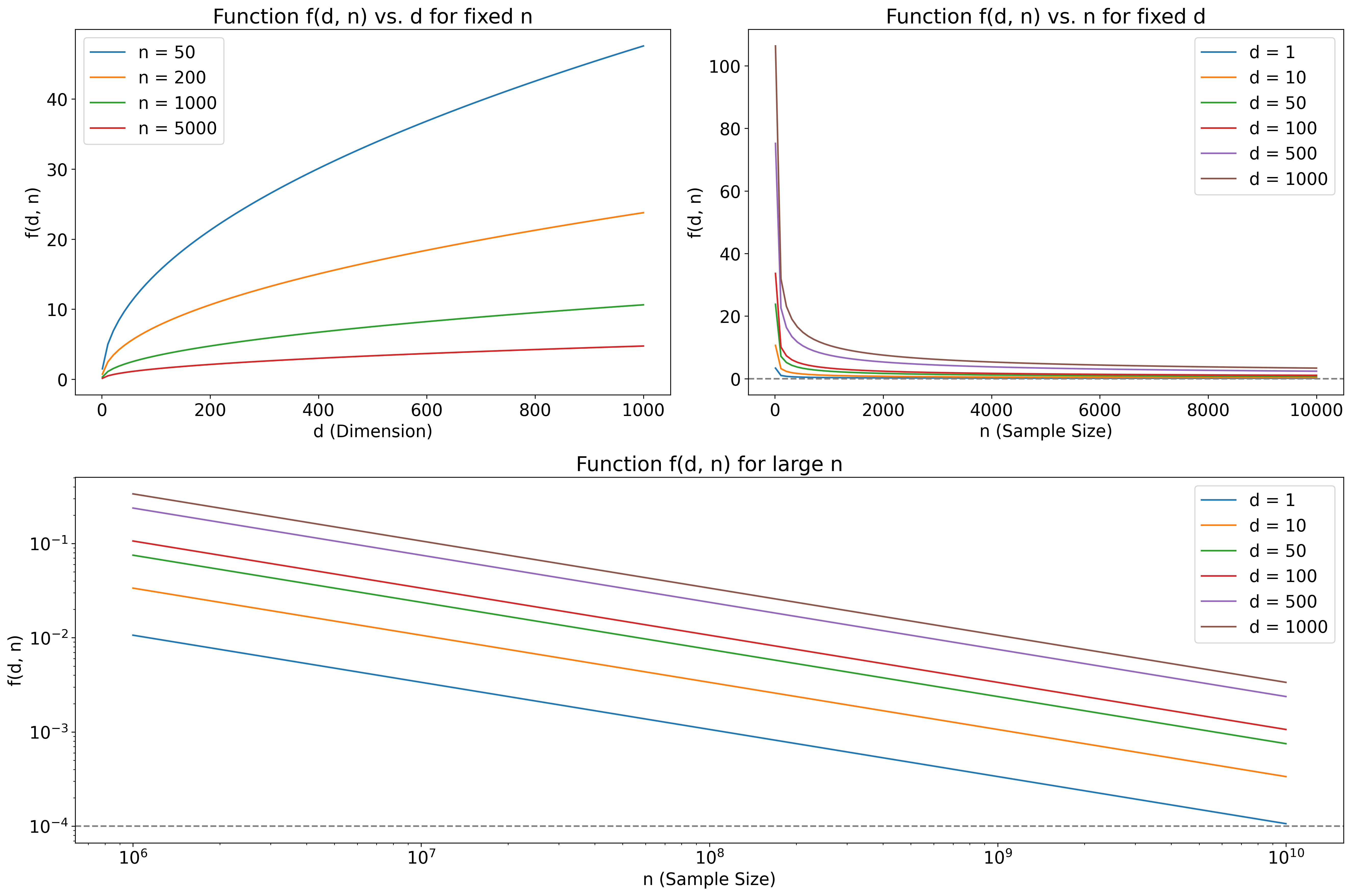}
    \caption{Rademacher Complexity $f(d,n)$ as a function of the number of samples $n$ and the dimensionality $d$. We have applied the $\log$ scaling on both axes for the third figure (second row) for better visualization.}
    \label{fig:RademacherComplexity}
\end{figure*}

In this section, we describe the main contribution of our work by providing a generalization bound for quantum machine learning models in noisy environments. Building upon statistical learning theory~\cite{vapnik1998statistical,mohri2018foundations} and quantum information theory~\cite{nielsen2002quantum}, we prove our main result in Appendix~[\ref{subsec:genbundApp}]:

    \begin{theorem}\label{th:gboundmain}
        [Generalization Bound for Parameterized Quantum Models] Let $d,N \in \mathbb{N}$ and $\delta \in [0,1)$. Consider a $d$-dimensional parameter space $\Theta \subset \mathbb{R}^d$ a class of quantum machine learning model functions $\mathcal{F}_\Theta = \{ f_{\theta,p}: \theta \in \Theta\}$ where each model output under a noisy channel with noise parameter $p \in [0,1$) is given by: $f_{\theta,p}(x) = \eta(p)f_\theta(x)$, with $\eta:\,[0,1)\to (0,\infty)$ a perturbation function satisfying $\eta(0)=1$. Assume the following:
        \begin{itemize}
            \item \textbf{Loss Function}: The loss l$:\mathcal{Y}\times\mathbb{R}\to[0,1]$ is Lipschitz continuous in its second argument with Lipschitz constant $L>0$.
            \item \textbf{Perturbed model bounded gradient}: The gradient of the perturbed model $f_{\theta,p}(x)$ w.r.t $\theta$ is bounded by the Lipschitz constant $L_f^p$:
            \(
            ||\nabla_\theta f_{\theta,p}(x)|| \leq L_f^p, \forall x \in \mathcal{X}, \theta \in \Theta.
            \)
            \item \textbf{FIM Lower Bound}: Let $\mathcal{F}(\theta)$ denote the quantum Fisher Information Matrix associated with the model. Suppose there exists $m>0$ such that:
            \[
            \sqrt{\text{det}(\mathcal{F}(\theta))} \ge m > 0, \quad \forall \theta \in \Theta.
            \]
        \end{itemize}
        Let $V_\Theta$ be the volume of the parameter space $\Theta$ and define:
        \begin{equation*}\label{eq:defineCprime}
            C^\prime = \log V_\Theta - \log V_d - \log m + d \log L_f^p ,
        \end{equation*}
        where $V_d = \frac{\pi^{\frac{d}{2}}}{\Gamma\left(\frac{d}{2} + 1\right)}$ is the volume of a unit ball in $\mathbb{R}^d$, and $\Gamma(\cdot)$ is the gamma function. Then, for any $\delta>0$, with probability at least $1-\delta$ over the random draw of an i.i.d. training set $D = \{(x_i,y_i)\}_{i=1}^N \sim P$, the following generalization bound holds uniformly for all $\theta \in \Theta$:
        \begin{equation}\label{eq:generalizationBound}
            R(\theta) \leq \hat{R}_N(\theta) + \frac{12\sqrt{\pi d} \exp(\frac{C^\prime}{d})}{\sqrt{N}} + 3\sqrt{\frac{\log(2/\delta)}{2N}}.
        \end{equation}
        where $R(\theta) = \mathbb{E}_{(x,y)\sim P} [l(y,f_{\theta,p}(x))]$ is the expected risk and $\hat{R}_N(\theta) = \frac{1}{N} \sum\limits_{i=1}^N l(y_i,f_{\theta,p}(x_i))$ is the empirical risk.
\end{theorem}

\begin{table*}
  \centering
  \begin{tabular}{|r|r|r|r|} \hline
      \textbf{d} & \textbf{k(d)} & \textbf{N} & \textbf{Third Term}, $\delta = 0.005$\\ 
      \hline
      1 & 2.72 & 8 & 0.99 \\
      10 & 3.49 & 13 & 0.60 \\
      100 & 10.10 & 102 & 0.07 \\
      1000 & 31.65 & 1002 & 0.01 \\
      10000 & 100.01 & 10002 & 0.00 \\
      50000 & 223.61 & 50002 & 0.00 \\
      100000 & 316.23 & 100002 & 0.00 \\

      \hline
  \end{tabular}
  \caption{Estimated generalization error scaling for increasing model dimensionality. Each column shows the parameter dimension $d$, a complexity measure $k(d)$, the approximate sample size $N$ required to reduce complexity-driven overhead, and the resulting complexity term of the bound. Higher dimensionality increases complexity, yet the growth is modest; with sufficient samples, even very high-dimensional models can achieve negligible complexity contributions.}
  \label{tab:GeneralizationError}
\end{table*}

Theorem~\ref{th:gboundmain} provides a data-dependent generalization bound for parameterized quantum models under noise. It reveals how sample size $N$, parameter dimension $d$, and model complexity interplay. For large $N$, the third term $\sim \sqrt{\frac{\log(2/\delta)}{N}}$ becomes negligible. Consequently, the dominant complexity term $\frac{12\sqrt{\pi d}\exp(C^{\prime}/d)}{\sqrt{N}}$ and the empirical risk $\hat{R}_N(\theta)$ determine the model’s capacity to generalize.

As $d$ grows large, $\exp(C^{\prime}/d) \to 1$, so the complexity scales approximately as $\sqrt{d}$. Intuitively, having more parameters can increase complexity, yet the \(\tfrac{1}{\sqrt{N}}\) factor ensures that as N grows, the model complexity effectively diminishes, improving generalization. Nevertheless, increasing the dimension $d$ can make generalization more challenging if the dataset is small.

This bound also emphasizes that, for any fixed $d$, the complexity term decreases as $N$ increases, suggesting that, in the limit $N\to\infty$, the model's generalization error can be made arbitrarily small. However,  as shown in Fig.~\ref{fig:RademacherComplexity} for any finite $N$, there is always a non-zero complexity term. Thus, even though we may achieve empirical risk close to zero, the interplay of dimension and sample size ensures that perfect generalization is not guaranteed.

To gain further insight, consider:
\[
k(d) = \sqrt{d}\,\exp\left(\frac{C^{\prime}}{d}\right).
\]
For large $d, k(d)$ grows slower than linearly but increases with d. Roughly, one requires on the order of $k(d)^2$ data points to push the complexity-driven error below a threshold of $12\sqrt{\pi}$. From table~\ref{tab:GeneralizationError}, it is apparent that by appropriately increasing the number of training samples $N$, one can significantly push down the complexity term, making it nearly minimal for large data sets. The practical challenge of generalizing with high-dimensional parameter spaces can be tackled by ensuring adequate sample sizes. This aligns with the well-established notion in machine learning that more data leads towards better generalization. While high-dimensional models might start more complex, the capacity to temper that complexity with larger datasets allows such models to remain viable, reinforcing that sufficient data can offset the dimensional burden.

The global bound in Theorem~\ref{th:gboundmain} considers the worst-case scenario over the entire parameter space $\Theta$. In practice, the learned parameter $\hat{\theta}$ may reside in a much smaller, effectively low-complexity region after training converges. This motivates local bounds that yield tighter guarantees.

\begin{corollary}\label{th1:corrlocalParams}[Local Generalization Bound]
Let the conditions of Theorem~\ref{th:gboundmain} hold. Suppose after training, $\hat{\theta}$ is a solution, and consider a local neighborhood $\Theta_{loc}\subset \Theta$ around $\hat{\theta} \text{ as } \Theta_{loc} := \{\theta \in \Theta : \|\theta - \hat{\theta}\|\leq \delta\}$ where $\delta>0$ but up to global limits. Assume within $\Theta_{loc}$:
\begin{itemize}
\item The Fisher Information Matrix remains well-conditioned: $\sqrt{\det(\mathcal{F}(\theta))} \ge m_{loc}>0$.
\item The gradients of the perturbed model are bounded locally by $L_{f_{loc}}^p$.
\end{itemize}
Define: $C_{loc} = \log(V_{\Theta_{loc}}) - \log(V_d) - \log(m_{loc}) + d \log(L_{f,loc}^p)$,
where $V_{\Theta_{loc}}$ is the volume of the local region. Then, with probability at least $1-\delta$, for all $\theta \in \Theta_{loc}$:
\begin{equation}\label{eq:localGeneralizationBound}
R(\theta) \leq \hat{R}N(\theta) + \frac{12\sqrt{\pi d},\exp(C_{loc}/d)}{\sqrt{N}} + 3\sqrt{\frac{\log(2/\delta)}{2N}}.
\end{equation}

\begin{proof}
This follows from Theorem~\ref{th:gboundmain} by restricting the parameter space to the smaller, better-conditioned subset $\Theta_{loc}$. Replacing global parameters $m$ and $L_f^p$ with their local counterparts $m_{loc}$ and $L_{f_{loc}}^p$, and using $V_{\Theta_{loc}}$ in place of $V_\Theta$, leads to the stated local bound. The same Rademacher complexity and covering number arguments apply over a reduced search space, potentially tightening the bound.
\end{proof}
\end{corollary}

Corollary~\ref{th1:corrlocalParams} shows that if training converges to a region where parameters vary less and the Fisher information remains stable, the complexity term decreases. This improved local conditioning can yield sharper generalization guarantees.

\begin{corollary}\label{th1:deffQFIM}[Effective Dimension and QFIM]
Let $\mathcal{F}(\theta)$ denote the QFIM and assume: $\sqrt{\det(\mathcal{F}(\theta))} \ge m > 0, \quad \forall \theta \in \Theta$. Suppose $\lambda_1(\theta)\geq \cdots \geq \lambda_d(\theta)>0$ are the eigenvalues of $\mathcal{F}(\theta)$. For a threshold $\alpha>0$, define the effective dimension: $d_{\mathrm{eff}}(\alpha)=\max\{r\leq d: \lambda_r(\theta)\geq \alpha,\ \forall \theta\in\Theta\}$. Then, applying the reasoning of Corollary~\ref{th1:corrlocalParams} to an effective dimension setting, we obtain with probability at least $1-\delta$:
\begin{equation}\label{th1:cordeff}
R(\theta) \leq \hat{R}N(\theta) + \frac{12 \sqrt{\pi d_{\mathrm{eff}}(\alpha)}\,\exp(C_{loc}/d_{\mathrm{eff}}(\alpha))}{\sqrt{N}} + 3\sqrt{\frac{\log(2/\delta)}{2N}}.
\end{equation}

\begin{proof}
By focusing on the subspace where the QFIM eigenvalues exceed $\alpha$, we effectively reduce the dimension to $d_{\mathrm{eff}}(\alpha)$. The argument parallels that of Corollary~\ref{th1:corrlocalParams}, simply substituting the effective dimension $d_{\mathrm{eff}}(\alpha)$ and corresponding constants into the main theorem's complexity estimates.
\end{proof}
\end{corollary}

Corollary~\ref{th1:deffQFIM} underscores that not all parameters are equally relevant for generalization. The effective dimension $d_{\mathrm{eff}}(\alpha)$ filters out parameter directions that do not significantly affect the state's distinguishability. This refinement can yield meaningful, dimension-reduced generalization bounds, bridging the gap between theoretical complexity measures and practical model performance.

Overall, Theorem~\ref{th:gboundmain} and its corollaries highlight how the interplay of sample size $N$, parameter dimension $d$, and the geometry of the QFIM govern the generalization capabilities of quantum machine learning models. The effective dimension perspective further refines these insights, guiding the design and training of models that achieve favorable tradeoffs between expressiveness and generalization.

\section{Numerical Analysis}\label{sec:numericalAnalysis}
Next, we discuss the numerical analysis to validate the proposed theoretical generalization bound. Our primary goal is to investigate how the global and local bounds behave under different noise levels, dataset complexities, and sample sizes. To that end, we designed experiments using a quantum neural network with two qubits, evaluated under depolarizing noise with rates $p \in \{0.05,0.1,0.5\}$. By varying $p$, we examine mild, moderate, and strong noise conditions, thus assessing the robustness of the bounds across a spectrum of realistic NISQ-era scenarios. These experiments were performed on two datasets of different complexities: (1) the Iris dataset, restricted to the first two classes for simplicity, and (2) the Digits (MNIST) dataset, focusing on digits $\{0,1\}$ with features reduced to $8$ dimensions via principal component analysis. Ref.~\cite{khanal2024generalization} informs that these datasets are real-world dataset representatives of quantum machine learning tasks for generalization analysis. These datasets are particularly relevant for the NISQ era as the Iris dataset is small and simple for low-dimensional analysis, while the MNIST dataset has higher inherent complexity but can be reduced to a manageable size for near-term quantum devices. We repeated each experiment three times to assess the variability of results across multiple runs. This approach allowed us to examine the bound's tightness, consistency, and applicability under noise conditions relevant to the NISQ era. 

\subsection{Methodology}

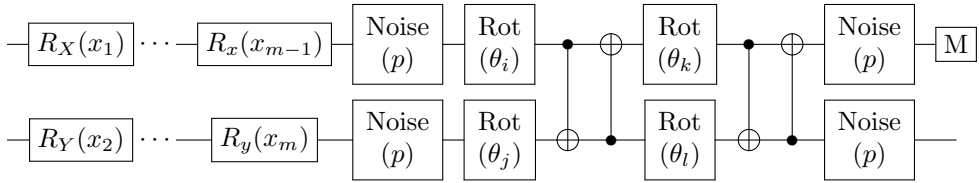
\begin{figure*}[ht]
    \centering
    
    \Qcircuit @C=0.8em @R=0.6em {
         & \gate{R_X(x_1)} & \cdots & & \gate{R_x(x_{m-1})} & \gate{\begin{array}{c} \text{Noise} \\ (p) \end{array}} & \gate{\begin{array}{c} \text{Rot} \\ (\theta_i) \end{array}} & \ctrl{1} & \targ & \gate{\begin{array}{c} \text{Rot} \\ (\theta_k)  \end{array}} & \ctrl{1} & \targ & \gate{\begin{array}{c} \text{Noise} \\ (p) \end{array}} & \gate{\text{M}} \\
         & \gate{R_Y(x_2)} & \cdots &  & \gate{R_y(x_{m})} &  \gate{\begin{array}{c} \text{Noise} \\ (p) \end{array}} & \gate{\begin{array}{c} \text{Rot} \\ (\theta_j)  \end{array}} & \targ & \ctrl{-1} & \gate{\begin{array}{c} \text{Rot} \\ (\theta_l) \end{array}} & \targ & \ctrl{-1} &  \gate{\begin{array}{c} \text{Noise} \\ (p) \end{array}} & \qw
        }
    \caption{Two-qubit circuit with depolarizing noise channels for noise rate $ p \in \{0.05,0.1,0.5\}$, parameterized single-qubit rotations, and controlled operations, preliminary CNOT gate. The Rot $(\cdot)$ is a single-qubit rotation gate with three Euler angles. $\theta_i,\theta_j,theta_k$ and $\theta_l$ are the parameters with distinct three Euler angles. The measurement is performed on the first qubit on a computation basis. }
    \label{fig:qccircuit}
    \end{figure*}

We employ a 2-qubit parameterized quantum circuit illustrated in Fig.~\ref{fig:qccircuit} to model the quantum machine learning task. While the same circuit structure can be applied to both datasets, more complex datasets might eventually require deeper or more sophisticated circuits for optimum performance. For a circuit with $n_{\text{layers}}$ layers and $n_{\text{qubits}}$ qubits, $d = n_{\text{layers}} \cdot n_{\text{qubits}} \cdot c$, where $c$ is a constant reflecting the number of parameters introduced per qubit per layer. For example, with $n_{\text{qubits}} = 2$ and $n_{\text{layers}} = 2$, and assuming $c = 3$ (e.g., for parameterized single-qubit rotations), we have $d = 12$. To ensure uniform bounds and consistency, we define the global parameter space as $\Theta = [-2\pi, 2\pi]^d$, resulting in a global volume of $(4\pi)^d$.

The model architecture consists of an embedding layer and a variational quantum circuit. Classical inputs $\mathbf{x} \in \mathbb{R}^m$ are mapped into quantum states via parameterized single-qubit rotations (e.g., $R_x(x_i)$ or $R_y(x_i)$). Noise is incorporated by inserting depolarizing channels with noise rate $p \in [0,1)$. The model then becomes $f_{\theta,p}(x) = \eta(p) f_\theta(x)$, where $\eta(p)$ scales the expectation value due to noise. This noise has a smoothing effect on the loss landscape, reducing gradient magnitudes and enhancing Fisher Information stability locally.

We train the model by minimizing mean square error loss using a natural gradient descent method. Gradient estimates are obtained using the parameter-shift rule. We trained the model for $20$ epochs and a fixed number of runs $(n_{\text{runs}}=3)$ to average the random initialization effects. After training, we defined a local neighborhood $\Theta_{\text{loc}}$ around $\hat{\theta}$ to ensure the stability of the quantum FIM. We choose a hypercube for computational simplicity, defined as $\{\theta: \|\theta - \hat{\theta}\|_{\infty} \leq \delta\}$, with a local volume of $(2\delta)^d$. The radius $\delta$ is determined by a continuity-based procedure, ensuring that the minimal eigenvalue or determinant-based quantity of the quantum FIM within $\Theta_{\text{loc}}$ remains above a fraction (e.g., $\alpha = 0.5$) of its value at $\hat{\theta}$. This ensures a stable neighborhood size and minimal conditioning related to the quantum FIM. The value of $\delta$ is also constrained to be within the global limits, for example, $ \alpha< \delta \leq 2\pi$.

Finally, we compute a local Lipschitz constant $L_{f,\text{loc}}^p$ by sampling gradients within $\Theta_{\text{loc}}$ and taking the maximum norm. By restricting complexity measures to these stable neighborhoods, the derived local bound becomes tighter and more representative of the model's actual generalization behavior.

\subsection{Result}\label{subsec:result}
\begin{figure*}
\centering
  \includegraphics[width=\linewidth]{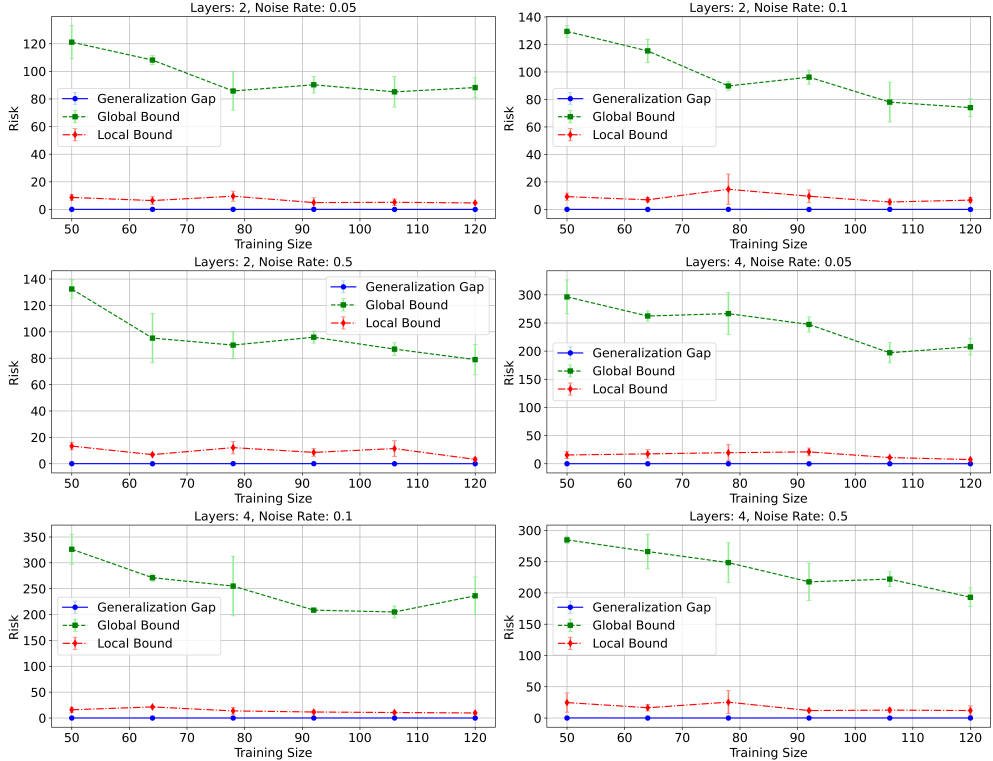}
  \caption{Global and local bounds versus the generalization gap for the first two classes of the Iris dataset. The local bound more closely tracks the observed gap, indicating a tighter and more accurate estimate of generalization.}
  \label{fig:resultiris}
  \end{figure*}
\begin{figure*}
\centering
  \centering
  \includegraphics[width=\linewidth]{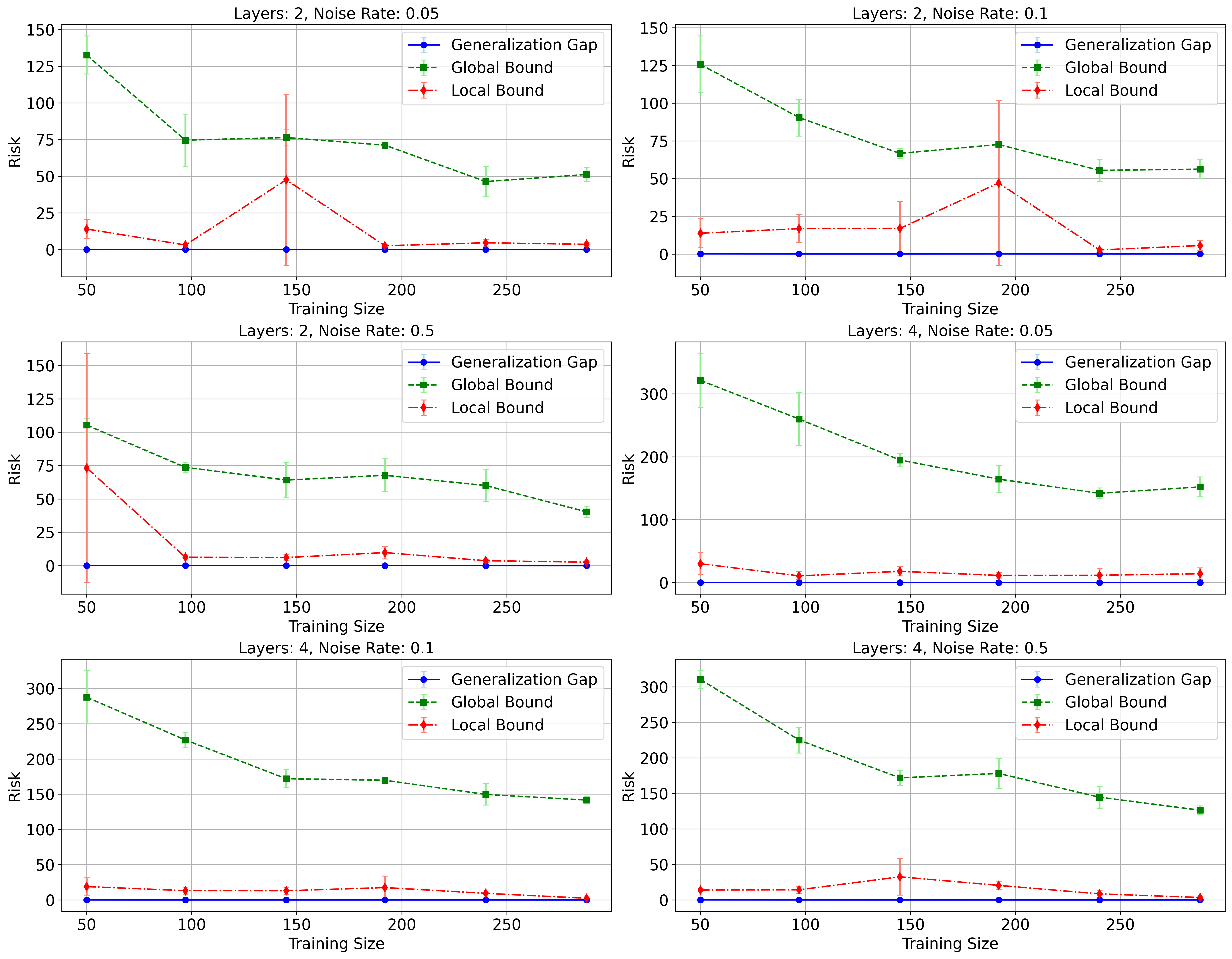}
  \caption{Global and local bounds compared to the observed generalization gap for MNIST digits $\{0,1\}$. The local bound shows occasional spikes, possibly due to random parameter initializations, but remains closer to the empirical gap than the global bound.}

  \label{fig:subfig2}
\label{fig:resultmnist}
\end{figure*}

For each dataset, we plot the generalization gap, global, and local bound across varying sample sizes, layers, and noise rates in Fig.~[\ref{fig:resultiris},\ref{fig:resultmnist}]. Each plot highlights the generalization gap, global bound, and local bound as a function of training size for different combinations of layers and noise rates. Error bars represent standard deviation over $3$ runs. The local bound's accuracy holds across different model depths, noise levels, and datasets. The result suggests that focusing on the local geometry around the trained parameters could get a more realistic and informative assessment of generalization performance. The generalization gap remains low across all configurations, and we observe that as the training size increases, the generalization gap tends to decrease or stay close to minimal values.

The global bound starts at a high level for the Iris dataset but decreases as the sample size increases. Nevertheless, it remains significantly above the observed generalization gap, implying a conservative nature typical of worst-case complexity arguments. In contrast, the local bound is consistently closer to the generalization gap, though not perfectly aligned. This improved proximity of the local bound suggests that refining complexity estimates to a stable neighborhood around $\hat{\theta}$ and yields more noteworthy upper estimates of the risk. The local region's constraints, influenced by stable Fisher Information conditions and locally computed Lipschitz constants, appear to capture the effective complexity more accurately.

On the Digits dataset, the global bound again stands high above the actual gap, but the local bound offers smaller and more stable estimates. Although not perfectly matching the gap, the local bound's relative closeness and a steadier downward trend as $N$ grows demonstrate that local complexity assessments adapt more naturally to increasing sample sizes. This adaptability supports the argument that local bounds are not merely a heuristic refinement but a necessary step toward a more faithful characterization of generalization in quantum models.

With the Iris dataset, where dimensionality and complexity are relatively small, global and local bounds descend as training grows, and the local bound maintains a more moderate scale. The Digits dataset experiments, involving larger input spaces and potentially more complex variations, show that the global bound still inflates above the actual gap. Yet, the local bound remains stable and moves closer to capturing the effective complexity of the learned model. This stability and reduced scale in the local bound, observed across multiple runs and conditions (varying noise rates and layer counts), supports our theorem~\ref{th:gboundmain} that local analysis provides a more accurate and data-dependent complexity assessment.

Despite differences in dimension and complexity, the stable performance of the same PQC on both Iris and Digits suggests that even a small QNN can learn binary classification tasks with varying complexity when suitably optimized. However, deeper or more sophisticated circuits might be needed for more complex versions of the Digits dataset or other larger-scale problems. Additionally, increasing the depolarizing noise rate $p$ leads to a narrower range of parameter-space fluctuations, tightening the local bound but also limiting model performance. This tradeoff between noise robustness and model performance is a common theme in quantum machine learning, and our results suggest that local complexity-based bounds can help navigate this tradeoff more effectively.

Thus, the local bound is conceptually appealing and empirically supported. The differences between global and local complexity trends hint that complexity should not be viewed uniformly across the entire parameter space. Instead, effective complexity is sensitive to the region around the trained parameters. Noise, model architecture, and available data samples interact to create a local landscape where complexity-based bounds align more with observed performance. A key finding is that global uniform bounds do not tightly reflect real-world performance when noise is present, creating a larger theoretical gap. By incorporating noise effects via the effective dimension, our analysis offers a more accurate account of how noise constrains the model’s capacity.

\section{Discussion}\label{sec:discussion}
Our approach to quantifying generalization in quantum machine learning places the geometry of parameterized quantum states at the center of understanding how noise and finite data affect predictive reliability. The parameter space volume, training sample size, and the QFIM emerge as central elements governing complexity. The QFIM, a measure of parameter distinguishability~\cite{helstrom1969quantum,braunstein1994statistical,paris2009quantum}, provides insights into how parameter variations relate to model sensitivity, which can inform stability under realistic noise conditions. Classical learning theory often relies on measures such as the VC dimension or the Rademacher complexity~\cite{vapnik1998statistical,bartlett2002rademacher}, yet those concepts lack a direct connection to the geometric and operator-theoretic properties of quantum states. Incorporating the QFIM into the analysis integrates geometric and statistical principles, producing a perspective that relates parameter scaling, data volume, and controlled complexity growth.

Our framework applies across various parameter dimensions and noise levels, and the complexity terms depend on volume measures and gradients within parameter space. The local neighborhood analysis uses the QFIM to focus on parameter regions where complexity remains stable. The effective dimension, derived from QFIM eigenvalues, provides a mechanism for identifying parameter directions that carry predictive significance. This approach extends beyond generic notions of large parameter counts or deep circuits, isolating meaningful subspaces that maintain manageable complexity. Such reasoning relates to efforts linking trainability to structural properties of parameterized circuits~\cite{mcclean2018barren,holmes2022connecting}, introducing a route for shaping model architectures guided by geometric criteria rather than expansions of parameter space.

The statistics collected from numerical experiments are consistent with the theoretical complexity scaling. Larger parameter dimension increases complexity in a predictable manner related to $\sqrt{d}$, and larger training sizes correspond to lower complexity-driven errors. These observations connect the theoretical bound directly to practical scenarios. The observed controlled complexity growth suggests that scaling parameter dimension in quantum models can lead to predictable generalization behavior, provided the interplay between dimension and data volume is managed. The results on effective dimension and local parameter neighborhoods yield tools that may influence the design of parameterized quantum circuits by directing attention toward stable parameter directions and conditions that maintain balanced complexity.

This line of reasoning extends beyond raw measures of expressiveness, linking the predictive reliability of trained quantum models to underlying geometric structures. Conventional arguments about quantum models often revolve around their large state spaces, yet without quantifying how parameter variations interact with data availability and noise. Well-defined complexity bounds clarify that quantum models can retain consistent behavior as problem sizes grow. Parameters no longer appear as an unstructured collection but as a network of directions differentiated by QFIM eigenvalues. Identifying and controlling these directions through appropriate training and architectural choices may inform research on QML capacity, optimization strategies, and representational limits~\cite{cerezo2021variational,larocca2023theory}. Complexity control is not an abstract ideal; it becomes tangible through geometric metrics and data-driven constraints.

By connecting the QFIM to generalization properties, our analysis relates the structural features of quantum states to the statistical aspects of learning. Instead of relying on heuristic claims or incomplete reasoning, the results rely on rigorously derived bound and direct empirical support. The outcome is a systematic view of complexity emergence and reduction, guiding theoretical analysis and practical implementations of quantum learning models.

\section{Conclusions}\label{sec:conclusion}
Our work utilizes geometric and statistical principles to relate quantum Fisher information, parameter space volume, and training data to the complexity of learning in noisy quantum models. The proposed bound and numerical results suggest that structured parameter subsets, identified through QFIM eigenvalues, provide a means to ensure stable generalization performance without uncontrolled complexity growth. This work provides a more nuanced understanding of complexity in quantum machine learning by explicitly linking it to the geometry of quantum states and the QFIM, complementing previous studies on expressiveness and capacity. While previous work has explored model capacity and expressiveness in quantum machine learning~\cite{havlivcek2019supervised,biamonte2017quantum,ciliberto2018quantum,schuld2019quantum,lloyd2013quantum}, this work introduces a framework that directly connects parameter-dependent complexity to the tangible geometry of quantum states via the QFIM.

From the perspective of the work proposed here, complexity emerges as a controllable quantity, motivating future research that refines parameter selection or training protocols to preserve stability in large-scale or resource-constrained learning tasks. Work on adaptive circuit structures, informed by local QFIM-based criteria, may help identify directions that maintain manageable complexity while improving representational fidelity. The proposed theorem incorporates noise effects through a general perturbation function, ensuring that performance remains stable under various conditions. Investigating individual or combined noise channels, diverse parametrization, or dynamic processes offers avenues for refining model architectures and optimizing strategies further to enhance adaptability and scalability within the QML context. The insights presented here may stimulate investigations into hybrid quantum-classical designs and application-tailored model architectures, paving the way for more robust, scalable, and interpretable quantum machine learning methods.

\section*{Declarations}
\bmhead{Funding} 
Part of this work was performed while P.R. was funded by the National Science Foundation under Grant Nos. 2136961, and 2210091, and 1905043. 
The views expressed herein are solely those of the author(s) and do not necessarily reflect those of the National Science Foundation.

\bmhead{Conflict of interest}
The authors have no relevant financial or non-financial interests to disclose.

\bmhead{Availability of data and materials}
Data sharing does not apply to this article as no datasets were generated or analyzed during the current study. This article is a systematic literature review, and as such, it synthesizes and analyzes information from previously published literature.

\begin{appendices}\label{Appendix}
\section{}
\subsection{Parameter Space Geometry}\label{app:A}
In the main text, we introduced the Fisher information matrix (FIM) $\mathcal{F}(\theta)$ as a Riemannian metric on the parameter space $\Theta \subset \mathbb{R}^d$. This induces a natural geometric structure on $\Theta$: the geodesic distance between $\theta$ and $\theta^{\prime}$ measures how different these parameter points are in terms of their influence on the model's predictions rather than just their Euclidean distance.

For a parameterized model and its associated FIM, we consider the volume element induced by $\mathcal{F}(\theta)$:
\begin{equation}\label{eq:pspaceVolume}
dV(\theta) = \sqrt{\det(\mathcal{F}(\theta))}\, d\theta^1 \cdots d\theta^d.
\end{equation}
A geodesic ball $B(\theta,\epsilon)$ of radius $\epsilon$ centered at $\theta$ has volume
\begin{equation}\label{eq:geodesicvolume}
V(\theta,\epsilon) = \int_{B(\theta,\epsilon)} dV(\theta^{\prime}) = \int_{B(\theta,\epsilon)} \sqrt{\det(\mathcal{F}(\theta^{\prime}))}\, d\theta^{\prime}. 
\end{equation}

If $\sqrt{\det(\mathcal{F}(\theta))}$ is bounded by a positive constant $m>0$, then for small $\epsilon$, we can approximate:
\begin{equation}\label{eq:geoballVolumeApproximate}
V(\theta,\epsilon) \approx V_d \epsilon^d \sqrt{\det(\mathcal{F}(\theta))} \geq V_d \epsilon^d m,
\end{equation}
where
\[
V_d = \frac{\pi^{d/2}}{\Gamma\left(\tfrac{d}{2} + 1\right)}
\]
is the volume of the unit ball in $\mathbb{R}^d$. Intuitively, a lower bound on $\sqrt{\det(\mathcal{F}(\theta))}$ ensures that geodesic balls are not "too small," implying fewer truly distinct parameter configurations at $\epsilon$. This geometric insight underpins the covering number bound we discuss next.

\subsection{Covering Numbers of a Parameter Space}
The covering number of a parameter space \(\Theta\) measures how many balls of radius \(\epsilon\) are required to cover the entire space. Since the FIM provides a natural measure of distinguishability, a lower bound on $\sqrt{\det(\mathcal{F}(\theta))}$ relates volumes of small balls to parameter distinguishability.
\begin{lemma}[Covering Number and Volume]\label{lemmaCoveringNumber}
Let $\Theta \subset \mathbb{R}^d$ be a compact parameter space with volume $V_\Theta$. Assume that the determinant of the Fisher Information Matrix $\mathcal{F}(\theta)$ satisfies:
\begin{equation*} \label{eq:detFIM}
    \sqrt{\det(\mathcal{F}(\theta))} \geq m > 0, \quad \forall \theta \in \Theta.
\end{equation*}
Then, for any $\epsilon > 0$, the covering number $N(\epsilon,\Theta,||\cdot||)$ of $\Theta$ with respect to the Euclidean norm $||\cdot||$ satisfies:
\begin{equation}\label{eq:coveringNumberVolumeRelationship}
    \log N(\epsilon,\Theta,||\cdot||) \leq C - d \log \epsilon,
\end{equation}
where $C = \log V_\Theta - \log V_d - \log m$ and $V_d = \frac{\pi^{\frac{d}{2}}}{\Gamma\left(\frac{d}{2} + 1\right)}$ is the volume of a unit ball in $\mathbb{R}^d$.

\begin{proof}
   Consider an $\epsilon$-ball around $\theta$:
\[
V(\theta,\epsilon) \geq V_d \epsilon^d m.
\]

    To cover $\Theta$, the total volume of $N$ such balls must be at least $V_\Theta$. So we have:
    \begin{equation}\label{eq:coveringVolume}
        N (\epsilon, \Theta, ||\cdot||)  \leq \frac{V_\Theta}{V_d \epsilon^d m}.
    \end{equation}
    Taking the logarithm:
\begin{align*}
    \log N(\epsilon, \Theta, | \cdot | ) &\leq \log V_\Theta - \log \left( V_d \epsilon^d m \right) \notag \\
     &= \log V_\Theta - \log V_d - \log m - d \log \epsilon \notag \\
    &= C - d \log \epsilon. 
\end{align*} 
This completes the proof of lemma~\eqref{lemmaCoveringNumber}. \end{proof}\end{lemma}
The inequality \eqref{eq:coveringNumberVolumeRelationship} shows how covering numbers scale with dimension d and resolution $\epsilon$. Smaller $\epsilon$ or larger d requires more balls, thus more complex model classes. Conversely, a larger m reduces complexity by ensuring a certain "geometric rigidity" in the parameter space. 

\section{}
\subsection{Bounding Empirical Rademacher Complexity}\label{sec:RademacherComplexityBound}
We now relate covering numbers to the empirical Rademacher complexity $\hat{\mathcal{R}}_N(\mathcal{F}_\Theta)$ of the model class $\mathcal{F}_\Theta = \{f_{\theta,p}:\theta\in\Theta\}$. The Rademacher complexity quantifies the model's capacity to fit random noise and thus provides an upper bound on generalization error.

\begin{lemma}[Empirical Rademacher Complexity Bound]\label{lemmaRademacher}
    Let $\mathcal{F}_\Theta = \{f_{\theta,p}: \theta \in \Theta\}$ be a model function class. Let $\mathcal{D} = \{x_i,y_i\}_{i=1}^N$ be a dataset of $N$ samples drawn from the distribution $P$. Assume that $f_{\theta,p}$ is bounded by the Lipschitz constant $L_f^p$ and the determinant of the quantum Fisher information matrix is at least $m$. Then the empirical Rademacher complexity of the model class $\mathcal{F}_\Theta$ is bounded by:
    \begin{equation}\label{eq:RademacherBoundfinal}
        \mathcal{\hat{R}}_N(\mathcal{F}_\Theta) \leq \frac{6 \sqrt{\pi d} \exp\left(\frac{C^\prime}{d} \right)}{\sqrt{N}}
    \end{equation}
where $C^\prime = \log V_\Theta - \log V_d - \log m + d \log L_f^p$.
\begin{proof}
    Given a dataset $\mathcal{D} = \{x_i,y_i\}_{i=1}^N$ the empirical Rademacher complexity of a model class $\mathcal{F}_\Theta$ is defined as:
    \begin{equation}\label{eq:Rademacher}
        \mathcal{\hat{R}}_N(\mathcal{F}_\Theta) = \mathbb{E}_{\sigma} \left[ \sup\limits_{f \in F_\Theta} \sum\limits_{i=1}^N \sigma_i f(x_i) \right],
    \end{equation}
    where $\sigma = \{\sigma_i\}_{i=1}^N$ are i.i.d Rademacher random variables. taking values in $\{-1,+1\}$ with equal probabilities. Dudley's entropy integral provides an upper bound on the empirical Rademacher complexity in terms of the covering numbers. If we let $\|\cdot\|_{2,\mathcal{D}}$ be the empirical $L_2$-norm over the sample $\mathcal{D}$, we have:
    \begin{equation}\label{eq:duleyRademacher}
        \hat{R}_N(\mathcal{F}_\Theta) \leq  \frac{12}{\sqrt{N}} \int_0^{\epsilon_{max}} \sqrt{\log N(\epsilon, \mathcal{F}_\Theta, ||\cdot||_{2,\mathcal{D}})} \, d\epsilon.
    \end{equation}
    where $N(\epsilon, \mathcal{F}_\Theta, ||\cdot||_{2,\mathcal{D}})$ is the covering number of $\mathcal{F}_\Theta$ with respect empirical $L_2$ metric: $||f - g ||_{2,\mathcal{D}} = \left(\frac{1}{N} \sum\limits_{i=1}^N (f(x_i) - g(x_i)) \right)^{\frac{1}{2}}$ and $\epsilon_{max} = \sup\limits_{f \in \mathcal{F}_\Theta} ||f||_{2,\mathcal{D}}$. Since $f_{\theta,p}$ is bounded by the Lipschitz constant $L_f^p$, we can write: $||f_{\theta,p} - f_{\theta^{\prime},p}||_{2,\mathcal{D}} \leq L_f^p ||\theta - \theta^{\prime}||, \quad \forall \theta, \theta^{\prime} \in \Theta$.
    This implies that an $\epsilon$-cover of $\Theta$ with respect to the Euclidean norm $||\cdot||$ is also an $\epsilon L_f^p$-cover of $\mathcal{F}_\Theta$ with respect to the empirical $L_2$ metric. Using Lemma~\ref{lemmaCoveringNumber}, we can write:
    \begin{equation*}
        \log N(\epsilon, \mathcal{F}_\Theta, ||\cdot||_{2,\mathcal{D}}) \leq \log N(\frac{\epsilon}{L_f^p}, \Theta, ||\cdot||) 
        \end{equation*}
        \begin{align}\label{eq:FunctionCovering}
        &\leq C - d \log\left( \frac{\epsilon}{L_f^p}\right) 
        = C - d \left( \log \epsilon - \log L_f^p \right) \notag\\
        &\implies\log N(\epsilon, \mathcal{F}_\Theta, ||\cdot||_{2,\mathcal{D}}) \leq C^\prime - d \log \epsilon, 
        \end{align} 

Substituting this result into into Eq.~\eqref{eq:duleyRademacher} gives:
\begin{equation}
    \hat{R}_N(\mathcal{F}_\Theta) \leq  \frac{12}{\sqrt{N}} \int_0^{\epsilon_{max}} \sqrt{C^\prime - d \log \epsilon} \, d\epsilon.
\end{equation}
We perform the elementary calculus in the following derivations, and if the reader trusts our calculations, they may skip the following steps.

Let $t = - \log \epsilon$, so $\epsilon = e^{-t}$ and $d\epsilon = -e^{-t} dt$. The limit changes as: when  $\epsilon = \epsilon_{max}, t = \log  \epsilon_{max}$ and as $\epsilon$ decreases from $\epsilon_{max}$ to 0, $t$ increases from $-\log \epsilon_{max}$ to 0. Thus:
\begin{align*}
    \hat{R}_N(\mathcal{F}_\Theta) &\leq  \frac{12}{\sqrt{N}} \int_{-\log \epsilon_{max}}^0 \sqrt{C^\prime - d \log e^{-t}} \, -e^{-t} dt \\
    \end{align*}
    Assuming $\epsilon_{max} = 1$ (since $f_{\theta,p}(x)$ is bounded in [0,1]), we have $t=0$ at $\epsilon = 1$, so:
    \begin{align*}
    \hat{R}_N(\mathcal{F}_\Theta) &\leq  \frac{12}{\sqrt{N}} \int_{0}^{\infty} \sqrt{C^\prime + d t} \, e^{-t} dt \\
    \end{align*}

    To make the derivation easier, let us define:
    \begin{equation}\label{eq:RademacherIntegral}
    I = \int_{0}^{\infty} \sqrt{C'+dt} e^{-t} dt
    \end{equation}
    Let $S = C' +dt$, then $t = \frac{S-C'}{d}$ and $\frac{dt}{dS} = \frac{d}{dS} \left( \frac{S - C'}{d} \right) \implies dt = \frac{dS}{d}$ and $e^{-t} = e^{-\left(\frac{S-C'}{d}\right)} = e^{\frac{C'}{d}}e^{\frac{-S}{d}}$. Since, $C' \text{ and } d$ are constants, we can write:
    \begin{equation*}
    I = \int_{C'}^{\infty} \sqrt{S} e^{\frac{C'}{d}} e^{\frac{-S}{d}} \frac{dS}{d} = \frac{e^{\frac{C'}{d}}}{d} \int_{C'}^{\infty} \sqrt{S} e^{\frac{-S}{d}} dS
    \end{equation*}
        Let $u = \frac{s}{d} \implies s = du \text{ and } ds = d du$. Thus,
    \begin{equation*}
    I = \frac{e^{\frac{C'}{d}}}{d} \int_{\frac{C'}{d}}^{\infty} \sqrt{du} e^{-u} d(du) 
    = e^{\frac{C'}{d}} \sqrt{d} \int_{\frac{C'}{d}}^{\infty} u^{\frac{1}{2}} e^{-u} du
    \end{equation*}
    This integral is the gamma function and can be written as:
    % \begin{equation}\label{eq:GammaFunction}
    $I = {e^{\frac{C'}{d}}}\sqrt{d} \Gamma\left(\frac{3}{2}, \frac{C'}{d}\right)$.
    % \end{equation}
    Since $\Gamma(s,a) \leq \Gamma(s)$ for $a > 0$, and $\Gamma\left( \frac{3}{2}\right) \ frac{\sqrt{\pi}}{2}$ we can write:
    \begin{equation}\label{eq:BoundedGamma}
    I \leq {e^{\frac{C'}{d}}}\sqrt{d} \frac{\sqrt{\pi}}{2}
    \end{equation}
    Substituting this back into Eq.~\eqref{eq:duleyRademacher}, we get:
    \begin{equation}\label{eq:RademacherBound}
    \hat{\mathcal{R}}_N(\mathcal{F}_\Theta) \leq \frac{6\sqrt{\pi d} \quad e^{\left(\frac{C'}{d}\right)}}{\sqrt{n}}
    \end{equation}
This completes the proof of the lemma~\eqref{lemmaRademacher}.
\end{proof}\end{lemma}
This lemma encapsulates how geometry (via quantum FIM) and parameter space volume control Rademacher complexity. As $N$ grows, the complexity term diminishes, indicating improved generalization potential.

Next, we discuss how the Rademacher complexity bound can be used to derive a generalization bound for quantum machine learning models.

\subsection{Generalization Bound}\label{subsec:genbundApp}
We now derive the generalization bound proposed in Theorem~\ref{th:gboundmain} in the main text. The bound follows from standard statistical learning theory and the Rademacher complexity result obtained above.
\begin{theorem}[Restatement of Theorem~\ref{th:gboundmain}]\label{th:main}
Let $d,N \in \mathbb{N}, \delta \in (0,1)$, and consider a parameter space $\Theta \subset \mathbb{R}^d$. The quantum model class $\mathcal{F}_\Theta=\{f{\theta,p}:\theta\in\Theta\}$ with noise parameter $p\in[0,1)$ satisfies $f_{\theta,p}(x)=\eta(p)f_\theta(x)$ with $\eta(0)=1$. Assume:
\begin{itemize}
\item The loss $l:\mathcal{Y}\times\mathbb{R}\to[0,1]$ is Lipschitz continuous in its second argument with constant $L \leq 1$.
\item The model gradients are bounded as: $\|\nabla_\theta f_{\theta,p}(x)\|\le L_f^p$.
\item The quantum FIM satisfies $\sqrt{\det(\mathcal{F}(\theta))}\ge m>0$.
\end{itemize}
Define
$
C^{\prime} = \log V_\Theta - \log V_d - \log m + d \log L_f^p$.

Then with probability at least 1-$\delta$ over an i.i.d. sample $D=\{x_i,y_i\}_{i=1}^N$ of size $N$,
\begin{equation}\label{eq:generalizationBoundApp}
R(\theta) \leq \hat{R}_N(\theta) + \frac{12\sqrt{\pi d}\exp(C^\prime/d)}{\sqrt{N}} + 3\sqrt{\frac{\log(2/\delta)}{2N}},
\end{equation}
uniformly for all $\theta \in \Theta$.

\begin{proof}
A standard result in learning theory states that with probability at least $1-\delta$:
\begin{equation}\label{eq:standardRademacher}
R(\theta) \leq \hat{R}_N(\theta) + 2\hat{\mathcal{R}}N(\mathcal{L}) + 3\sqrt{\frac{\log(2/\delta)}{2N}},
\end{equation}

where $\mathcal{L}=\{(x,y)\mapsto l(y,f_{\theta,p}(x)):\theta\in\Theta\}$.

Since $l(y,\hat{y})$ is Lipschitz with constant $L$ and $f_{\theta,p}(x)$ is bounded by $L_f^p$, it follows that $\mathcal{L}$ has Rademacher complexity at most $L\cdot L_f^p$. Applying Lemma~\ref{lemmaRademacher}:
\[
\hat{\mathcal{R}}_N(\mathcal{L}) \leq L L_f^p \frac{6\sqrt{\pi d}\text{ exp}{\left(C^{\prime}/d\right)}}{\sqrt{N}}.
\]
For $L\le 1$, this simplifies directly to:
\[
\hat{\mathcal{R}}_N(\mathcal{L}) \leq \frac{6\sqrt{\pi d}exp{\left(C^{\prime}/d\right)}}{\sqrt{N}}.
\]

Substitute back into the generalization inequality:
\begin{equation*}
R(\theta) \leq \hat{R}_N(\theta) + 2\cdot\frac{6\sqrt{\pi d}\text{ exp}{\left(C^{\prime}/d\right)}}{\sqrt{N}} + 3\sqrt{\frac{\log(2/\delta)}{2N}}.
\end{equation*}

This gives:
\begin{equation*}
R(\theta) \leq \hat{R}_N(\theta) + \frac{12\sqrt{\pi d}\text{ exp}{\left(C^{\prime}/d\right)}}{\sqrt{N}} + 3\sqrt{\frac{\log(2/\delta)}{2N}}, 
\end{equation*}
matching the statement of Theorem~\ref{th:gboundmain}.
\end{proof}
\end{theorem}

This completes the proof of the generalization bound for theorem~\ref{th:gboundmain}. The interplay between Fisher information geometry, covering numbers, and Rademacher complexity provides a path from parameter space properties to explicit generalization guarantees. In practice, after training, restricting to local regions or reducing dimension to the effective dimension often yields sharper bounds.

\end{appendices}

\bibliography{main}

\end{document}